\pdfoutput=1
\documentclass[11pt]{article}

\usepackage{acl}

\usepackage{times}
\usepackage{latexsym}

\usepackage[T1]{fontenc}
\usepackage[utf8]{inputenc}

\usepackage{microtype}

\usepackage{inconsolata}

\usepackage{graphicx}

\usepackage{amsmath,amsthm,amssymb,amsfonts}
\usepackage{booktabs}
\usepackage{multirow}
\usepackage{enumerate}
\usepackage{enumitem}
\usepackage{graphicx}
\usepackage{subfig}
\usepackage{epigraph}

\usepackage{pgfplots}
\pgfplotsset{compat=1.18}
\usepackage{pgfplotstable}
\usepackage{makecell}
\usepackage{color}
\usepackage{colortbl}  %彩色表格需要加载的宏包
\usepackage{xcolor}
\usepackage{array} 
\title{Query Routing for Homogeneous Tools: An Instantiation in the RAG Scenario}

\author{
 \textbf{Feiteng Mu\textsuperscript{1}\thanks{Work done during internship at Alibaba Inc.}},
 \textbf{Yong Jiang\textsuperscript{2}\thanks{Corresponding author.}},
 \textbf{Liwen Zhang\textsuperscript{2}},
 \textbf{Chu Liu\textsuperscript{2}},
\\
 \textbf{Wenjie Li\textsuperscript{1}},
 \textbf{Pengjun Xie\textsuperscript{2}},
 \textbf{Fei Huang\textsuperscript{2}},
\\
 \textsuperscript{1}The Department of Computing, The Hong Kong Polytechnic University, Hong Kong\\
 \textsuperscript{2}Institute for Intelligent Computing, Alibaba Group\\
\texttt{\{csfmu,cswjli\}@comp.polyu.edu.hk,\{yongjiang.jy,zlw439616\}@alibaba-inc.com}
}

\begin{document}
\maketitle

\begin{abstract}
% Current research on tool learning focuses on selecting the most effective tool from a large number of options, without adequately considering cost-effectiveness which is an important criterion in human problem-solving. In this paper, we investigate the problem of selection of homogeneous tools. 
% For homogeneous tools that can solve the specific task, we predict both their performance and the cost required to achieve the task. 
% We then cost-effectively assign queries to the optimal tools on demand.
% The experiment demonstrates that our method can achieve higher performance with a lower cost compared to strong baselines.

Current research on tool learning primarily focuses on selecting the most effective tool from a wide array of options, often overlooking cost-effectiveness, a crucial factor in human problem-solving. In this paper, we address query routing for homogeneous tools by predicting both their performance and the associated cost required to accomplish a given task. We then assign queries to the optimal tools in a cost-effective manner. Our experimental results demonstrate that our method achieves higher performance at a lower cost compared to strong baseline approaches.

\end{abstract}

"\textit{Don't use a sledgehammer to crack a nut.}" \\
\rightline{~~------~~\textit{Proverb}}

\section{Introduction}

% Despite their breathtaking capabilities \cite{feng2024improving,zhang2024benchmarking}, large language models (LLMs) \cite{touvron2023llama,bai2023qwen,achiam2023gpt} often present plausible yet factually incorrect or outdated information  \cite{ji2023survey,zhang2023hallucination} to user queries.
% % (often referred to as hallucination)
% % schick2024toolformer
% To alleviate this issue, tool learning has become a crucial strategy \cite{qin2023tool,qu2024tool}, which enables LLMs to interact with the real world.

Tool learning \cite{qin2023tool,qu2024tool} aims to arm large language models (LLMs) \cite{touvron2023llama,bai2023qwen,achiam2023gpt} with real world tools, to alleviate hallucinations \cite{ji2023survey,zhang2023hallucination} of LLMs.

\begin{figure}
    \centering
    \includegraphics[width=\linewidth]{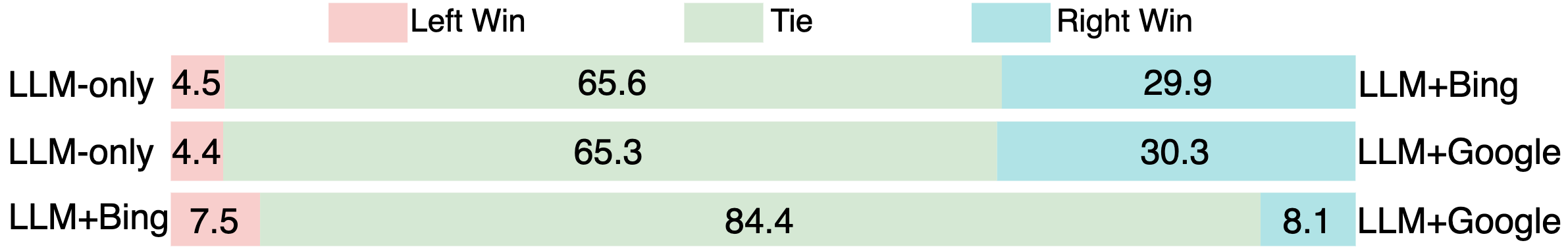}
    \caption{The win/tie/lose rates on our PrivateTimeQA test set when the LLM uses two compared tools. Besides Bing and Google, we consider a non-retrieval baseline, and denote the method as "LLM-only".}
    \label{fig:win_tie_lose}
\end{figure}

Existing tool learning methods focus on routing a query to the most effective tool from a large number of options \cite{qin2023toolllm,tang2023toolalpaca}, 
while overlooking the crucial aspect of cost-effectiveness, which is a significant criterion in human problem-solving.
% without adequately considering cost-effectiveness which is an important criterion in human problem-solving.
To bridge this gap, query routing for homogeneous tools has become an important issue. Taking the retrieval-augmented generation (RAG) \cite{lewis2020retrieval,gao2023retrieval} scenario as an example.
Given a user query, a RAG system first uses a search tool to retrieve the background documents. Then, a LLM reads the query and documents to give the response.
Given the availability of various web search tools, such as Google and Bing, the choice of which tool to use for a given query becomes an important consideration. 
First, a query that cannot be solved by one tool might be resolved by another. As shown in Figure \ref{fig:win_tie_lose}, Our dataset indicates that (1) in 4.5\% of cases, directly utilizing the LLM to answer queries surpasses the performance of the Bing-based RAG. And (2) there are 8.1\% of instances where the Bing-based RAG fails to deliver a solution, whereas the Google-based RAG successfully provides one.
Furthermore, different tools come with varying usage costs. From a cost-effectiveness perspective, it is preferable to select the least expensive tool that can effectively solve the query.
If we can adaptively allocate queries to the optimal tool, we can achieve a balance between cost and performance. Additionally, leveraging the complementarity between tools offers the potential for enhanced overall performance.
The key challenge in achieving this goal is estimating the performance of candidate tools in handling user queries, instead of making decisions after calling all tools for user queries \cite{chen2023frugalgpt,jiang2023llm}.
To overcome this challenge, we automatically construct training data including queries submitted to all tools, along with the returned scores. The training data allows us to learn neural models to predict the performance that LLM calls each tool to solve each query.

In this work, we explore query routing for homogeneous tools and instantiate it in the RAG scenario. We propose a router that assigns input queries to the most appropriate search tool without directly accessing any of the tools. To achieve this, we develop an automated pipeline to construct training data without manual effort. 
% Additionally, we introduce a label-refinement strategy to mitigate noise in labels caused by imperfect information retrieval processes. 
Finally, we design various assignment strategies to allocate user queries to the optimal tool as needed.

In summary, we make the following contributions.
\textbf{(1)} We propose a general approach that dynamically assigns an input query to the optimal tool from a set of homogeneous candidates, only based on the input query. 
\textbf{(2)} As far as we know, we are the first to consider the selection of homogeneous tools. Our framework is generalizable to any type of homogeneous tools, extending beyond the RAG scenario. 
\textbf{(3)} We evaluate our approach on several QA tasks with multiple LLMs. Experiments show that our method outperforms the use of a single, fixed search tool across different datasets.

\section{Related Work}
% 每个任务有自己的工具。没考虑到同一个任务可能有多个工具，但是这些工具有tradeoff。如何自适应的选择调用哪个？

\paragraph{Tool Learning}
Existing tool-learning methods focus on developing different tools to solve various problems. \cite{schick2024toolformer} teaches LLMs to call tools like Calculator, Calendar, etc.
\cite{qin2023toolllm} facilitates LLMs to master 16000+ real-world APIs.
\cite{zheng2024toolrerank,qu2024colt,du2024anytool} select the most suitable tool for a given query from a vast tool set, according to the similarity between the query and tool descriptions.
Differently, we investigate the problem of selecting homogeneous tools, which are less considered.

\paragraph{LLMs Selection}
Earlier, Different LLMs charged differently due to their uneven performance.  Practitioners focus on effectively choosing LLM to save costs.
\cite{chen2023frugalgpt} gradually employs more expensive LMs until a satisfactory performance is obtained. 
\cite{lu2023routing,vsakota2024fly,shnitzer2023large} propose the neural routing function that can precisely distribute each query to the expert LLMs.
% \cite{liu2024optllm} develops a multi-objective optimization method that maximizes the percentage of queries processed accurately, as well as minimizes the total cost of invoking LLMs.
However, recently, the cost of using LLMs has continued to decrease. 
The usage cost of tools has become a major obstacle to LLMs-based applications.
Thus, we explore the selection of homogeneous tools to achieve a balance between performance and cost.

\section{Method}
\subsection{Problem Formulation}

We demonstrate our work within the RAG framework to provide a realistic illustration. It should be noted that our approach is universal and applicable to various scenarios, not limited to just RAG.
Suppose that a LLM needs to solve a set of user queries. The LLM is permitted to call a search tool to generate responses.
We consider a situation where there are $M$ homogeneous tools $\{T_m\}_{m=1}^M$, e.g., Bing and Google, each with different costs. Our goal is to develop an adaptive assignment method to achieve a cost-performance trade-off.

% The key to achieving this is to estimate the probability that each tool will successfully address each query.

\begin{figure}[!tb]
    \centering
    \includegraphics[width=0.95\linewidth]{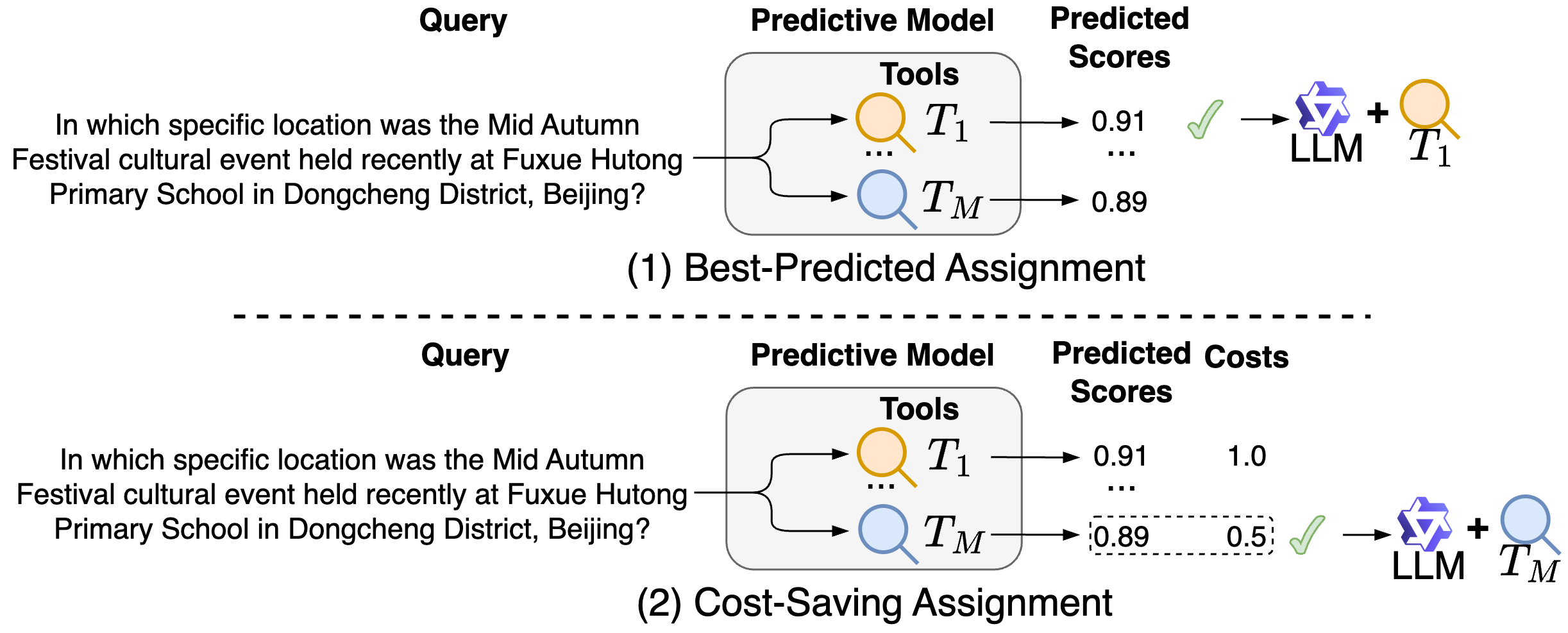}
    \caption{The framework of our method.
    We first predict the scores where the LLM calls each tool to solve each query. Then, we design different strategies to assign each query to the optimal tool on demand.}
    \label{fig:overall_framwork}
\end{figure}

We decompose this process into two steps. In the first step, we learn a predictive model $\mathcal{M}$ that predicts the scores where the LLM calls each tool to solve each query. 
We restrict $\mathcal{M}$ to only depend on the input query $q$ and the used tool $T_m$:
\begin{equation}
\setlength{\belowdisplayskip}{4pt} 
\setlength{\abovedisplayskip}{4pt}
\small
    p_{m}= \mathcal{M}(q,T_m), \forall m\in\{1,\cdots,M\}.
\end{equation}
In the second step, we devise diverse strategies to assign a query to the optimal tool on demand. The framework of our method is shown in Figure \ref{fig:overall_framwork}.

Next, we introduce the training of $\mathcal{M}$ and the implementation of the assignment strategies.

\subsection{Training Predicative Model}
\label{sec:training_predictor}
% Identifying the impact of tools on a particular query is difficult, as it remains uncertain until the tool actually processes the input. 
% To address this issue, we assume that the training data, including logged queries previously submitted to all tools, along with the returned responses by LLMs, are available, which allows us to train neural models that can be generalized to unseen queries. 

\paragraph{Data Preparation} We develop an automated pipeline to construct training data. Given the observed data $\mathcal{D}$ consisting of query-answer pairs $\{(q_n,r_n^{gold})\}_{n=1}^{|\mathcal{D}|}$, for each $q_n$, the LLM calls each search tool to obtain the query-related documents, and generates retrieval-augmented response $r_n^m$:
\begin{equation}
\setlength{\belowdisplayskip}{4pt} 
\setlength{\abovedisplayskip}{4pt}
\small
    \begin{split}
        \text{doc}_{n}^m&= T_m(q_n),\\
        r_{n}^m&= LLM(q_n, \text{doc}_{n}^m).
    \end{split}
\end{equation}
Then, we use the Text-Matching score between the generated answer $r_{n}^m$ and the ground-truth answer $r_n^{gold}$ to automatically calculate the labeled score when the LLM calls $T_m$ to handle $q_n$:
% can be  automatically calculated by:
\begin{equation}
\setlength{\belowdisplayskip}{4pt} 
\setlength{\abovedisplayskip}{4pt}
\small
    l_{n}^m= \text{Text-Matching}(r_{n}^m, r_n^{gold}).
\end{equation}
% where the metric is set as \textit{accuracy}.

\paragraph{Training} 
Given collected $\{(q_n,\{l_{n}^m\}_{m=1}^M)\}_{n=1}^{|\mathcal{D}|}$, 
we train $\mathcal{M}$ through the following objective:
\begin{equation}
\setlength{\belowdisplayskip}{6pt} 
\setlength{\abovedisplayskip}{6pt}
\small
    \begin{split}
    p_n^m&= \mathcal{M}(q_{n},T_m),\\
        \min_{\mathcal{M}} & \frac{1}{|\mathcal{D}|} \sum_{n=1}^{|\mathcal{D}|} \sum_{m=1}^{M} \mathcal{L}(p_n^m, l_{n}^m),
    \end{split}
    \label{eq:training_router}
\end{equation}
where we set $\mathcal{L}$ as Mean Square Error and initialize $\mathcal{M}$ as RoBERTa \cite{liu2019roberta}. The objective directly simulates the score of the LLM calling each tool to solve each query, enabling it to handle unseen cases where only the query is available.

\subsection{Assignment Strategies}
\label{sec:assign_methods}
Given search tools $\{T_m\}_{m=1}^M$, we denote their usage costs as $\{C_m\}_{m=1}^M$.
Assuming the user has a collection of queries $\{q_n\}_{n=1}^\mathcal{N}$ that need to be processed, we first use $\mathcal{M}$ to obtain the predicted scores $\{p_{n}^m\}_{m=1}^M$ for the query $q_n$. Then, inspired by \cite{lu2023routing,vsakota2024fly,shnitzer2023large}, we consider the following strategies to assign queries to corresponding tools.

(1) \textbf{Fixed-Tool}: LLM always calls a fixed tool to handle queries.
(2) \textbf{Best-Performance}: for $q_n$ and the predicted $\{p_{n}^m\}_{m=1}^M$, the LLM selects $T_{m^\star}$ $(m^\star=\mathop{\arg\max}_m p_{n}^m)$ with the maximal predicted score to handle $q_n$.
% (3) Threshold: LLM selects the cheapest tool among tools whose performance, i.e., the predicted $p_{nm}$, exceeds a user-defined threshold.
(3) \textbf{Cost-Saving}: This strategy aims to ensure that the average predicted score, i.e., $p_{n}^m$, exceeds a user-defined threshold while minimizing the average cost.
This problem can be solved by the integer linear programming (ILP):
% \begin{equation}
% \small
% \setlength{\belowdisplayskip}{4pt} 
% \setlength{\abovedisplayskip}{4pt}
% \begin{array}{ccc}
%  \text{minimize} & \sum w_{n}^m C_{m} \\
%  s.t. & \frac{1}{\mathcal{N}}\sum_{n=1}^{\mathcal{N}} \sum_{m=1}^M w_{n}^m p_{n}^m \geq P_{min} \\
%   & \sum_{m=1}^M w_{n}^m == 1, \forall n\in \{1,\cdots,\mathcal{N}\}
% \end{array}
% \end{equation}

\begin{equation}
\small
\setlength{\abovedisplayskip}{-4pt}
\setlength{\belowdisplayskip}{6pt} 
\begin{aligned}
    \text{minimize} \quad & \sum w_{n}^m C_m \\
    \text{s.t.} \quad & \frac{1}{\mathcal{N}} \sum_{n=1}^{\mathcal{N}} \sum_{m=1}^{M} w_{n}^m p_{n}^m \geq P_{\text{min}} \\
                     & \sum_{m=1}^{M} w_{n}^m = 1, \quad \forall n \in \{1, \cdots, \mathcal{N}\},
\end{aligned}
\end{equation}
where the variable $w_{n}^m \in \{0,1\}$ indicates  whether to assign $q_n$ to $T_m$, $P_{\text{min}}$ denotes the threshold.

% We collect and annotate an internal dataset from one anonymous E-commerce
% website. The dataset contains 25 named entity
% labels for goods in short texts. We also collect
% 300,000 unlabeled sentences for semi-supervised
% training.

\section{Experiment}
\subsection{Experimental Setup}

\paragraph{Datasets and Training details}
We focus on the QA scenario. 
The adopted datasets include public datasets CDQA \cite{xu2024cfresh} and WebQA \cite{li2016webqazh}, as well as a private dataset PrivateTimeQA\footnote{We leave the detail in Appendix \ref{app:statistcs}.} that we construct ourselves.
Data processing details, dataset statistics, and training details are shown in Appendix \ref{app:statistcs}.

% The adopted datasets include  public datasets, e.g., HotpotQA \cite{yang-etal-2018-hotpotqa}, FreshQA \cite{vu2023freshllms}, CDQA \cite{xu2024cfresh}, WebQA \cite{li2016webqazh} and a private dataset, TimeQA\footnote{We leave the detail in Appendix \ref{app:statistcs}.}, that we built ourselves.
% Data processing details, dataset statistics, and training details are shown in Appendix \ref{app:statistcs}.

\paragraph{Baselines} We use multiple LLMs, including Qwen-Max\footnote{https://dashscope.aliyuncs.com/compatible-mode/v1} \cite{bai2023qwen}, ChatGPT, and GPT4 in all of our experiments, due to their strong reasoning ability.
The used search tools include \textbf{Quark}, \textbf{Bing}, and \textbf{Google}.
Considering that the LLM can answer easy queries independently, we set a \textbf{non-retrieval} baseline, i.e., the LLM directly responds to the query. Consequently, its usage cost is set to 0. 
In addition, since existing methods \cite{shnitzer2023large,vsakota2024fly} assign queries to the most suitable LLM, we devise "LLMs-UB", Specifically,
For a query and corresponding scores obtained from several non-retrieval methods, such as Qwen-only, ChatGPT-only, and GPT4-only, we always assign the query to the LLM with the highest score. Therefore, this can be seen as an \textbf{u}pper \textbf{b}ound of \cite{shnitzer2023large,vsakota2024fly}.

\subsection{Evaluation the Predictive Model}

We first examine the ability of $\mathcal{M}$ to simulate the score that LLM calls each tool to solve test queries.

\subsubsection{Metrics}
For the testset $\{(q_n, \{l_{n}^m\}_{m=1}^{M})\}_{n=1}^\mathcal{N}$, 
we report the final QA \textit{accuracy} by selecting the tool with the maximal predicted $p_{n}^m$ for RAG:
\begin{equation}
\setlength{\belowdisplayskip}{6pt} 
\setlength{\abovedisplayskip}{6pt}
\small
    \begin{split}
    m^\star &=\mathop{\arg\max}_m p_{n}^m, \forall n\in \{1,\cdots,\mathcal{N}\},\\
        \text{accuracy}&=\frac{1}{\mathcal{N}} \sum_{n=1}^{\mathcal{N}} l_{n}^{m^\star}.
    \end{split}
\end{equation}
We also report the average cost of each method. For simplicity, we set the average costs of Quark, Bing, and Google to 0.33, 2, and 1, respectively.
 
% We also use the Kullback-Leibler (KL) distance of distributions between the predicted $p_{n}^m$ and the real $l_{n}^m$ as a metric, since a lower distance means a more precise simulation:
% \begin{equation}
% \setlength{\belowdisplayskip}{4pt} 
% \setlength{\abovedisplayskip}{4pt}
% \small
%     \mathcal{J}=\frac{1}{\mathcal{N}}\sum_{n=1}^{\mathcal{N}} \operatorname{KL}[\text{softmax}_m (p_{n}^m)|| \text{softmax}_m(l_{n}^m)].
% \end{equation}

% since a lower distance means a more accurate simulation.

% \paragraph{Baselines} We devise following baselines. (1) \textit{\textbf{Random}} which randomly selects a tool. (2) \textit{\textbf{Fixed-Tool}} which selects a fixed tool. Since we consider Bing Search, Google Search, as well as a non-retrieval condition, we denote the combinations between the LLM and tools as "LLM-only", "LLM+Bing", "LLM+Google". (3) The ablated variant \textit{\textbf{w/o Refinement}} that ablates the label-refinement module. It also select the tool with the maximal predicted $p_n^m$.

\begin{table}[!htb]
\centering\small
\setlength{\tabcolsep}{2pt}
\begin{tabular}{@{}lllllll@{}}
\toprule
\multirow{2}{*}{Methods} & \multicolumn{2}{c}{\tiny{PrivateTimeQA}} & \multicolumn{2}{c}{CDQA} & \multicolumn{2}{l}{WebQA} \\ \cmidrule(l){2-7} 
 & Acc. & Cost & Acc. & Cost & Acc. & Cost \\ \midrule
 
\rowcolor{gray!10}
\multicolumn{7}{c}{\textit{ChatGPT}} \\
LLM-only & 30.47 & 0 & 29.91 & 0 & 59.25 & 0 \\
LLMs-UB & 55.64 & 0 & 48.32 & 0 & 93.00 & 0 \\
LLM+Quark & 67.81 & 0.33 & 55.91 & 0.33 & 93.47 & 0.33 \\
LLM+Bing & 66.63 & 2 & 60.08 & 2 & 91.18 & 2 \\
LLM+Google & 68.63 & 1 & 61.13 & 1 & \textbf{94.07} & 1 \\
Ours (Roberta-large) & \textbf{68.89} & 1.24 & \textbf{63.28} & 0.68 & 93.55 & 0.52 \\ \midrule
\rowcolor{gray!10}
\multicolumn{7}{c}{\textit{GPT4}} \\
LLM-only & 29.9 & 0 & 28.46 & 0 & 71.77 & 0 \\
LLMs-UB & 55.64 & 0 & 48.32 & 0 & 93.00 & 0 \\
LLM+Quark & 71.87 & 0.33 & 57.22 & 0.33 & 94.42 & 0.33 \\
LLM+Bing & 70.66 & 2 & 61.22 & 2 & 93.36 & 2 \\
LLM+Google & 73.35 & 1 & 64.42 & 1 & \textbf{95.5} & 1 \\
Ours (Roberta-large) & \textbf{73.6} & 1.01 & \textbf{65.74} & 0.58 & 94.54 & 0.79 \\ \midrule
\rowcolor{gray!10}
\multicolumn{7}{c}{\textit{Qwen-max}} \\
LLM-only & 50.30 & 0 & 45.90 & 0 & 91.14 & 0 \\
LLMs-UB & 55.64 & 0 & 48.32 & 0 & 93.00 & 0 \\
LLM+Quark & 77.54 & 0.33 & 59.84 & 0.33 & \textbf{96.33} & 0.33 \\
LLM+Bing & 76.91 & 2 & 63.86 & 2 & 93.48 & 2 \\
LLM+Google & 77.95 & 1 & 66.75 & 1 & 94.97 & 1 \\
Ours (Roberta-large) & \textbf{78.41} & 1.23 & \textbf{69.56} & 1.20 & 94.68 & 1.30 \\ \bottomrule
\end{tabular}
\caption{The result of our method when we use ChatGPT, GPT4 and Qwen-max for response generation. "LLM-only" means the non-retrieval baseline. "LLM+Quark", "LLM+Bing" and "LLM+Google" means the LLM always calls Quark, Bing, or Google to solve queries. ``Acc. (\%)" denotes \textit{accuracy}. Scores with \textbf{bold} denote the best values among baselines.}
\label{tab:qa_overall}
\end{table}

% In addition, we develop the ablated variant to investigate the effectiveness of label refinement.

\begin{table}[!htb]
\centering\small
\setlength{\tabcolsep}{2pt}
\begin{tabular}{@{}llllll@{}}
\toprule
\multirow{2}{*}{Methods} & \multicolumn{1}{c}{\multirow{2}{*}{\begin{tabular}[c]{@{}c@{}}Model\\ Size\end{tabular}}} & \multicolumn{2}{c}{TimeQA} & \multicolumn{2}{c}{CDQA} \\ \cmidrule(l){3-6} 
 & \multicolumn{1}{c}{} & Acc. & Cost & Acc. & Cost \\ \midrule
\rowcolor{gray!10}
\multicolumn{6}{c}{\textit{ChatGPT}} \\
Ours (Roberta-base) & 125M & 68.68 & 1.24 & 62.3 & 0.68 \\
Ours (Roberta-large) & 355M & \textbf{68.89} & 1.24 & \textbf{63.28} & 0.68 \\
Ours (Qwen-0.5B) & 0.5B & 67.99 & 1.25 & 62.14 & 1.38 \\
Ours (Qwen-1.8B) & 1.8B & 68.73 & 1.02 & 61.37 & 1.03 \\ \midrule
\rowcolor{gray!10}
\multicolumn{6}{c}{\textit{GPT4}} \\
Ours (Roberta-base) & 125M & 72.48 & 1.45 & 65.57 & 0.83 \\
Ours (Roberta-large) & 355M & \textbf{73.6} & 1.01 & \textbf{65.74} & 0.58 \\
Ours (Qwen-0.5B) & 0.5B & 72.11 & 1.32 & 63.16 & 1.42 \\
Ours (Qwen-1.8B) & 1.8B & 73.06 & 1.10 & 64.76 & 1.07 \\ \bottomrule
\end{tabular}

\caption{The result of different backbones. }
\label{tab:backbones}
\end{table}

% \begin{table}[!htb]
% \centering\small
% \setlength{\tabcolsep}{2pt}
% \begin{tabular}{@{}llllll@{}}
% \toprule
% \multirow{2}{*}{Methods} & \multicolumn{1}{c}{\multirow{2}{*}{\begin{tabular}[c]{@{}c@{}}Model\\ Size\end{tabular}}} & \multicolumn{2}{c}{TimeQA} & \multicolumn{2}{c}{CDQA} \\ \cmidrule(l){3-6} 
%  & \multicolumn{1}{c}{} & Acc. & Cost & Acc. & Cost \\ \midrule
% GPT4-only & N/A & 29.9 & 0 & 28.46 & 0 \\
% LLMs-UB & N/A & 55.64 & 0 & 48.32 & 0 \\
% GPT4+Quark & N/A & 71.87 & 0.33 & 57.22 & 0.33 \\
% GPT4+Bing & N/A & 70.66 & 2 & 61.22 & 2 \\
% GPT4+Google & N/A & 73.35 & 1 & 64.42 & 1 \\ \midrule
% Ours (Roberta-base) & 125M & 72.48 & 1.45 & 65.57 & 0.83 \\
% Ours (Roberta-base) & 355M & \textbf{73.6} & 1.01 & \textbf{65.74} & 0.58 \\
% Ours (Qwen-0.5B) & 0.5B & 72.11 & 1.32 & 63.16 & 1.42 \\
% Ours (Qwen-1.8B) & 1.8B & 73.06 & 1.10 & 64.76 & 1.07 \\ \bottomrule
% \end{tabular}

% \caption{The result of predictive model evaluation. "LLM-only" means the non-retrieval baseline in which the LLM directly responds to the query. "LLM+Bing" and "LLM+Google" means the LLM always call Bing or Google search to solve queries. ``Acc. (\%)" denotes \textit{accuracy}. Scores with \textbf{Bold} denote the best values. }
% \label{tab:acc_cost_gpt4}
% \end{table}

\subsubsection{Results}
The result is shown in Table \ref{tab:qa_overall}. 
Whether using Qwen-max, ChatGPT, or GPT4 for response generation, LLM+Google usually outperforms LLM+Bing, LLM+Quark, and LLM-only, which coincides with human intuition that Google is currently the best search tool in the world. 
RAG outperforms LLM-only by a large margin, proving the necessity of utilizing tools for response generation. In addition, we have the following observations.

% \begin{itemize}[leftmargin=10pt, itemindent=0pt,itemsep=0pt,topsep=0pt,parsep=0pt]
    
% \end{itemize}

\textbf{(1)} On PrivateTimeQA and CDQA, our method significantly outperforms the best baseline, i.e., "LLM+Google". This trend is observed across multiple LLMs.
This is because despite "LLM+Google" achieving the highest overall accuracy, it still fails to solve some cases that "LLM+Quark" or "LLM+Bing" can handle. Specifically, the proportion of such samples is 6.6\% in the CDQA testset. 
Our method can adaptively assign these queries to the optimal tools, leading to better overall results. 
\textbf{(2)} Compared with "LLM+Bing", our method achieves a better result with a lower cost. Compared with "LLM+Google", our method has a better performance with a approaching cost. This demonstrates the generality of our method for different search engines and LLMs.

\subsection{Evaluation of the Adaptively Assignment}
\label{sec:cost_acc_ilp}
% \begin{figure}
%     \centering
%     \includegraphics[width=\linewidth]{figures/ilp_merged.png}
%     \caption{The cost-accuracy curve on merged testsets.}
%     \label{fig:merged_pc_curve}
% \end{figure}

\paragraph{Settings}
According to different strategies in Section \ref{sec:assign_methods}, we assign test queries to the corresponding tools and calculate the average accuracy and usage cost\footnote{Bing Search charges \$10 for 1000 requests, and Google Search charges \$5 for 1000 requests. }.
Cost–accuracy curves are plotted, which present the relationship between the average accuracy and the average cost per query needed. 

\begin{figure}[!ht]
  \centering
  % \subfloat[][]{\includegraphics[width=.8\linewidth]{figures/ilp_merged.png}}\quad
  \subfloat[The curve on the CDQA testset.]{\includegraphics[width=0.8\linewidth]{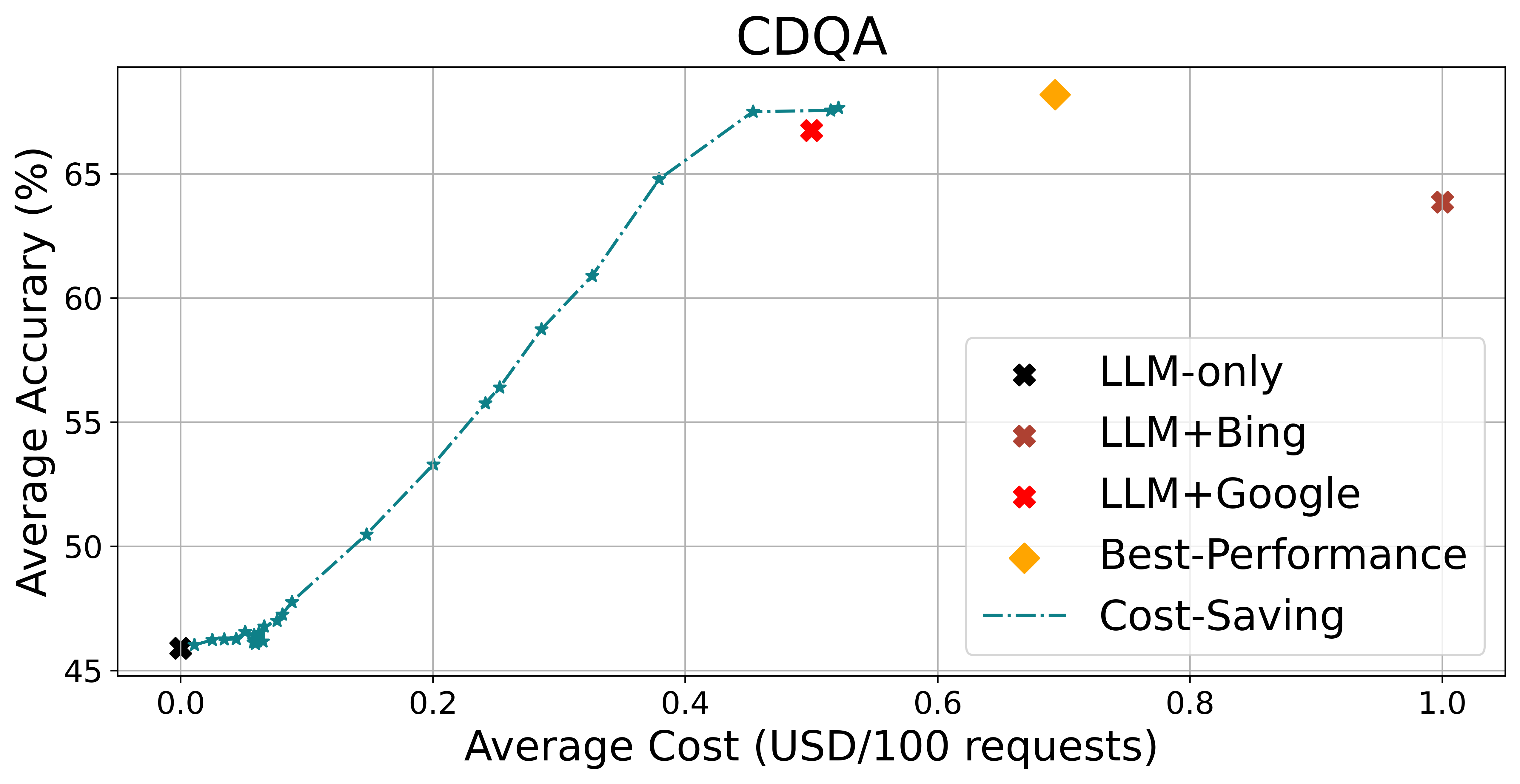}}\\
  % \subfloat[][]{\includegraphics[width=.45\linewidth]{figures/ilp_fourset.png}}\quad
  \subfloat[The curve on the PrivateTimeQA testset.]{\includegraphics[width=0.8\linewidth]{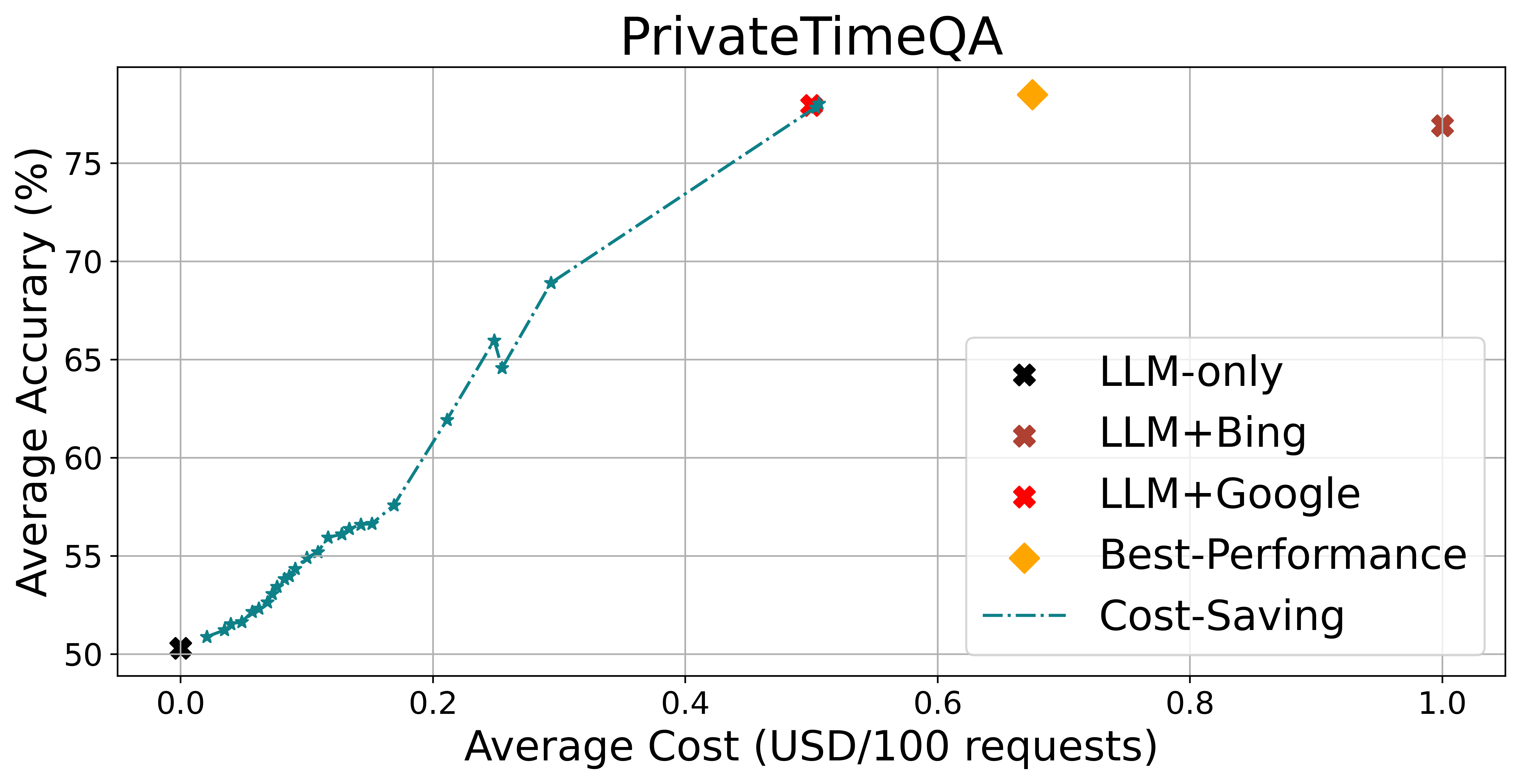}}
  \caption{The cost-accuracy curves on PrivateTimeQA and CDQA when using Qwen-max.}
  \label{fig:cfresh_time_pc_curve}
\end{figure}

\paragraph{Result}
We plot cost-accuracy curves on the test sets of PrivateTimeQA and CDQA in Figure \ref{fig:cfresh_time_pc_curve}.
"LLM+Google" has a higher accuracy and a lower cost compared to "LLM+Bing". We additionally have the following observations.
\textbf{(1)} Based on our $\mathcal{M}$, "Best-Performance"  achieves a higher accuracy than "LLM+Google", albeit at a higher cost. 
The accuracy increment comes from the supplement from "LLM+Bing", with a side-effect of higher cost.
\textbf{(2)} Compared to both "LLM+Google" and "Best-Performance", the "Cost-Saving" strategy is able to achieve higher accuracy with a lower cost. This shows that "Cost-Saving" is more flexible and can effectively save costs.

% \begin{figure}[!htb]
%     \centering
%     \includegraphics[width=\linewidth]{ilp_merged.png}
%     \caption{The cost-accuracy curve on the merged testset.}
%     \label{fig:ilp_merged}
% \end{figure}

% We additionally present the cost-accuracy curves on the testsets of CDQA and TimeQA. The result is shown in Figure \ref{fig:cfresh_time_pc_curve}. We have the findings that coincide with Section \ref{sec:cost_acc_ilp}.

\subsection{Further Discussion}
\paragraph{Impact of Different Backbones} 
We investigate the impact of different backbones for training our router. We consider Roberta-base(125M), Roberta-large(355M), Qwen-0.5B (0.5B), Qwen-1.8B(1.8B) for experiments. 
When using Roberta-large as the backbone, the result is better than that based on Roberta-base. 
But when using the larger Qwen-0.5B (0.5B), Qwen-1.8B (1.8B), even if we have tried various training techniques, the results do not increase significantly. We speculate that the decoder-only models are not suitable for this routing task which requires accurate numerical predictions. In addition, the larger model leads to higher latency. Therefore, we finally chose Roberta-large as our backbone. Due to limited computing resources, we are unable to attempt larger models such as Qwen-7b, Llama7b, etc.

% \begin{figure}
%     \centering
%     \includegraphics[width=0.95\linewidth]{alpha_balance.png}
%     \caption{Impact of $\alpha$ in the label refinement module.}
%     \label{fig:alpha_impact_chset}
% \end{figure}

% \begin{table}[!htb]
% \small
% \centering
% \setlength{\tabcolsep}{4pt}
% \begin{tabular}{@{}llll@{}}
% \toprule
%  & LLM-only & LLM+Bing & LLM+Google \\ \midrule
% Raw label $l_{n}^m$ & 0.71 & 0.71 & 0.57 \\
% Refined label $\widetilde{l_{n}^m}$ & 0.62 & 0.72 & 0.63 \\ \bottomrule
% \end{tabular}
% \caption{Case study for the label refinement.}
% \label{tab:casestudy_refine}
% \end{table}

% \paragraph{Case Analysis}
% We take a case to show why the label-refinement strategy works. For the query from HotpotQA: ``\textit{Where is the film studio located that produced the 1954 film The Egyptian?"}, we present its raw labels and refined labels obtained through different tools. As shown in Table \ref{tab:casestudy_refine}, 
% due to the noisy retrieval, the raw label obtained from "LLM+Google" are even smaller than "LLM-only". But with the label refinement, the refined label from "LLM+Google" gets larger than that from "LLM-only". In other words, the label refinement allows us to obtain a more robust partial order among multiple homogeneous tools.

\section{Conclusion}
In this work, we investigate the problem of selecting homogeneous tools for the sake of cost-effective trade-offs. We provide a clear definition of the problem and propose a general framework that can be adapted to any type of homogeneous tools.
The experiment shows that our method is effective and can achieve higher performance while at a lower cost compared to strong baselines. 
This indicates the potential of our framework where multiple homogeneous tools are available.

% The results demonstrate that our method not only achieves higher performance but also operates at a lower cost compared to strong baselines.

% \clearpage

\section*{Limitations}
Due to the high cost of search tools, our training data only contains approximately 16000 examples,. In the future, with a sufficient budget, we will conduct testing on more datasets. 
We use small encoding models to train our predictive model instead of LLMs, mainly for the following two reasons: (1) Tool selection has a high requirement for latency, while the speed of large models is relatively slow and does not meet the requirement. (2) Tool selection requires the model to predict the score of each tool solving each problem, which is actually a numerical prediction problem. However, the auto-regressive LLMs have significant issues in numerical prediction. In our experiment, the performance of LLMs was inferior to that of small encoding models.

% \section*{Ethical Considerations}

% In the used datasets, HotpotQA, FreshQA, WebQA, and CDQA are public datasets.
% TimeQA is a private dataset collected from public QA websites, which operates in the field of open-domain QA. 
% The dataset does not contain any sensitive, harmful, unethical, or personal information, but only open-domain questions and answers.
% We believe that our research work meets the ethics of ARR.

% \clearpage

% Bibliography entries for the entire Anthology, followed by custom entries
%\bibliography{anthology,custom}
% Custom bibliography entries only
\bibliography{anthology}

\begin{thebibliography}{23}
\providecommand{\natexlab}[1]{#1}

\bibitem[{Achiam et~al.(2023)Achiam, Adler, Agarwal, Ahmad, Akkaya, Aleman,
  Almeida, Altenschmidt, Altman, Anadkat et~al.}]{achiam2023gpt}
Josh Achiam, Steven Adler, Sandhini Agarwal, Lama Ahmad, Ilge Akkaya,
  Florencia~Leoni Aleman, Diogo Almeida, Janko Altenschmidt, Sam Altman,
  Shyamal Anadkat, et~al. 2023.
\newblock Gpt-4 technical report.
\newblock \emph{arXiv preprint arXiv:2303.08774}.

\bibitem[{Bai et~al.(2023)Bai, Bai, Chu, Cui, Dang, Deng, Fan, Ge, Han, Huang
  et~al.}]{bai2023qwen}
Jinze Bai, Shuai Bai, Yunfei Chu, Zeyu Cui, Kai Dang, Xiaodong Deng, Yang Fan,
  Wenbin Ge, Yu~Han, Fei Huang, et~al. 2023.
\newblock Qwen technical report.
\newblock \emph{arXiv preprint arXiv:2309.16609}.

\bibitem[{Chen et~al.(2023)Chen, Zaharia, and Zou}]{chen2023frugalgpt}
Lingjiao Chen, Matei Zaharia, and James Zou. 2023.
\newblock Frugalgpt: How to use large language models while reducing cost and
  improving performance.
\newblock \emph{arXiv preprint arXiv:2305.05176}.

\bibitem[{Du et~al.(2024)Du, Wei, and Zhang}]{du2024anytool}
Yu~Du, Fangyun Wei, and Hongyang Zhang. 2024.
\newblock Anytool: Self-reflective, hierarchical agents for large-scale api
  calls.
\newblock \emph{arXiv preprint arXiv:2402.04253}.

\bibitem[{Gao et~al.(2023)Gao, Xiong, Gao, Jia, Pan, Bi, Dai, Sun, and
  Wang}]{gao2023retrieval}
Yunfan Gao, Yun Xiong, Xinyu Gao, Kangxiang Jia, Jinliu Pan, Yuxi Bi, Yi~Dai,
  Jiawei Sun, and Haofen Wang. 2023.
\newblock Retrieval-augmented generation for large language models: A survey.
\newblock \emph{arXiv preprint arXiv:2312.10997}.

\bibitem[{Ji et~al.(2023)Ji, Lee, Frieske, Yu, Su, Xu, Ishii, Bang, Madotto,
  and Fung}]{ji2023survey}
Ziwei Ji, Nayeon Lee, Rita Frieske, Tiezheng Yu, Dan Su, Yan Xu, Etsuko Ishii,
  Ye~Jin Bang, Andrea Madotto, and Pascale Fung. 2023.
\newblock Survey of hallucination in natural language generation.
\newblock \emph{ACM Computing Surveys}, 55(12):1--38.

\bibitem[{Jiang et~al.(2023)Jiang, Ren, and Lin}]{jiang2023llm}
Dongfu Jiang, Xiang Ren, and Bill~Yuchen Lin. 2023.
\newblock Llm-blender: Ensembling large language models with pairwise ranking
  and generative fusion.
\newblock \emph{arXiv preprint arXiv:2306.02561}.

\bibitem[{Lewis et~al.(2020)Lewis, Perez, Piktus, Petroni, Karpukhin, Goyal,
  K{\"u}ttler, Lewis, Yih, Rockt{\"a}schel et~al.}]{lewis2020retrieval}
Patrick Lewis, Ethan Perez, Aleksandra Piktus, Fabio Petroni, Vladimir
  Karpukhin, Naman Goyal, Heinrich K{\"u}ttler, Mike Lewis, Wen-tau Yih, Tim
  Rockt{\"a}schel, et~al. 2020.
\newblock Retrieval-augmented generation for knowledge-intensive nlp tasks.
\newblock \emph{Advances in Neural Information Processing Systems},
  33:9459--9474.

\bibitem[{Li et~al.(2016)Li, Li, He, Wang, Cao, Zhou, and Xu}]{li2016webqazh}
Peng Li, Wei Li, Zhen He, Xuguang Wang, Ying Cao, Jie Zhou, and Wei Xu. 2016.
\newblock \href {https://api.semanticscholar.org/CorpusID:6901603} {Dataset and
  neural recurrent sequence labeling model for open-domain factoid question
  answering}.
\newblock \emph{ArXiv}, abs/1607.06275.

\bibitem[{Liu et~al.(2019)Liu, Ott, Goyal, Du, Joshi, Chen, Levy, Lewis,
  Zettlemoyer, and Stoyanov}]{liu2019roberta}
Yinhan Liu, Myle Ott, Naman Goyal, Jingfei Du, Mandar Joshi, Danqi Chen, Omer
  Levy, Mike Lewis, Luke Zettlemoyer, and Veselin Stoyanov. 2019.
\newblock \href {https://api.semanticscholar.org/CorpusID:198953378} {Roberta:
  A robustly optimized bert pretraining approach}.
\newblock \emph{ArXiv}, abs/1907.11692.

\bibitem[{Lu et~al.(2023)Lu, Yuan, Lin, Lin, Yuan, Zhou, and
  Zhou}]{lu2023routing}
Keming Lu, Hongyi Yuan, Runji Lin, Junyang Lin, Zheng Yuan, Chang Zhou, and
  Jingren Zhou. 2023.
\newblock Routing to the expert: Efficient reward-guided ensemble of large
  language models.
\newblock \emph{arXiv preprint arXiv:2311.08692}.

\bibitem[{Qin et~al.(2023{\natexlab{a}})Qin, Hu, Lin, Chen, Ding, Cui, Zeng,
  Huang, Xiao, Han et~al.}]{qin2023tool}
Yujia Qin, Shengding Hu, Yankai Lin, Weize Chen, Ning Ding, Ganqu Cui, Zheni
  Zeng, Yufei Huang, Chaojun Xiao, Chi Han, et~al. 2023{\natexlab{a}}.
\newblock Tool learning with foundation models.
\newblock \emph{arXiv preprint arXiv:2304.08354}.

\bibitem[{Qin et~al.(2023{\natexlab{b}})Qin, Liang, Ye, Zhu, Yan, Lu, Lin,
  Cong, Tang, Qian et~al.}]{qin2023toolllm}
Yujia Qin, Shihao Liang, Yining Ye, Kunlun Zhu, Lan Yan, Yaxi Lu, Yankai Lin,
  Xin Cong, Xiangru Tang, Bill Qian, et~al. 2023{\natexlab{b}}.
\newblock Toolllm: Facilitating large language models to master 16000+
  real-world apis.
\newblock \emph{arXiv preprint arXiv:2307.16789}.

\bibitem[{Qu et~al.(2024{\natexlab{a}})Qu, Dai, Wei, Cai, Wang, Yin, Xu, and
  Wen}]{qu2024colt}
Changle Qu, Sunhao Dai, Xiaochi Wei, Hengyi Cai, Shuaiqiang Wang, Dawei Yin,
  Jun Xu, and Ji-Rong Wen. 2024{\natexlab{a}}.
\newblock Colt: Towards completeness-oriented tool retrieval for large language
  models.
\newblock \emph{arXiv preprint arXiv:2405.16089}.

\bibitem[{Qu et~al.(2024{\natexlab{b}})Qu, Dai, Wei, Cai, Wang, Yin, Xu, and
  Wen}]{qu2024tool}
Changle Qu, Sunhao Dai, Xiaochi Wei, Hengyi Cai, Shuaiqiang Wang, Dawei Yin,
  Jun Xu, and Ji-Rong Wen. 2024{\natexlab{b}}.
\newblock Tool learning with large language models: A survey.
\newblock \emph{arXiv preprint arXiv:2405.17935}.

\bibitem[{{\v{S}}akota et~al.(2024){\v{S}}akota, Peyrard, and
  West}]{vsakota2024fly}
Marija {\v{S}}akota, Maxime Peyrard, and Robert West. 2024.
\newblock Fly-swat or cannon? cost-effective language model choice via
  meta-modeling.
\newblock In \emph{Proceedings of the 17th ACM International Conference on Web
  Search and Data Mining}, pages 606--615.

\bibitem[{Schick et~al.(2024)Schick, Dwivedi-Yu, Dess{\`\i}, Raileanu, Lomeli,
  Hambro, Zettlemoyer, Cancedda, and Scialom}]{schick2024toolformer}
Timo Schick, Jane Dwivedi-Yu, Roberto Dess{\`\i}, Roberta Raileanu, Maria
  Lomeli, Eric Hambro, Luke Zettlemoyer, Nicola Cancedda, and Thomas Scialom.
  2024.
\newblock Toolformer: Language models can teach themselves to use tools.
\newblock \emph{Advances in Neural Information Processing Systems}, 36.

\bibitem[{Shnitzer et~al.(2023)Shnitzer, Ou, Silva, Soule, Sun, Solomon,
  Thompson, and Yurochkin}]{shnitzer2023large}
Tal Shnitzer, Anthony Ou, M{\'\i}rian Silva, Kate Soule, Yuekai Sun, Justin
  Solomon, Neil Thompson, and Mikhail Yurochkin. 2023.
\newblock Large language model routing with benchmark datasets.
\newblock \emph{arXiv preprint arXiv:2309.15789}.

\bibitem[{Tang et~al.(2023)Tang, Deng, Lin, Han, Liang, and
  Sun}]{tang2023toolalpaca}
Qiaoyu Tang, Ziliang Deng, Hongyu Lin, Xianpei Han, Qiao Liang, and Le~Sun.
  2023.
\newblock Toolalpaca: Generalized tool learning for language models with 3000
  simulated cases.
\newblock \emph{arXiv preprint arXiv:2306.05301}.

\bibitem[{Touvron et~al.(2023)Touvron, Lavril, Izacard, Martinet, Lachaux,
  Lacroix, Rozi{\`e}re, Goyal, Hambro, Azhar et~al.}]{touvron2023llama}
Hugo Touvron, Thibaut Lavril, Gautier Izacard, Xavier Martinet, Marie-Anne
  Lachaux, Timoth{\'e}e Lacroix, Baptiste Rozi{\`e}re, Naman Goyal, Eric
  Hambro, Faisal Azhar, et~al. 2023.
\newblock Llama: Open and efficient foundation language models.
\newblock \emph{arXiv preprint arXiv:2302.13971}.

\bibitem[{Xu et~al.(2024)Xu, Li, Ding, Wang, Chen, Jiang, Deng, Ma, Zheng, Lu,
  Xie, Zhou, and Huang}]{xu2024cfresh}
Zhikun Xu, Yinghui Li, Ruixue Ding, Xinyu Wang, Boli Chen, Yong Jiang, Xiaodong
  Deng, Jianxin Ma, Hai-Tao Zheng, Wenlian Lu, Pengjun Xie, Chang Zhou, and Fei
  Huang. 2024.
\newblock \href {https://api.semanticscholar.org/CorpusID:268063789} {Let llms
  take on the latest challenges! a chinese dynamic question answering
  benchmark}.
\newblock \emph{ArXiv}, abs/2402.19248.

\bibitem[{Zhang et~al.(2023)Zhang, Li, Cui, Cai, Liu, Fu, Huang, Zhao, Zhang,
  Chen, Wang, Luu, Bi, Shi, and Shi}]{zhang2023hallucination}
Yue Zhang, Yafu Li, Leyang Cui, Deng Cai, Lemao Liu, Tingchen Fu, Xinting
  Huang, Enbo Zhao, Yu~Zhang, Yulong Chen, Longyue Wang, Anh~Tuan Luu, Wei Bi,
  Freda Shi, and Shuming Shi. 2023.
\newblock Siren's song in the ai ocean: A survey on hallucination in large
  language models.
\newblock \emph{arXiv preprint arXiv:2309.01219}.

\bibitem[{Zheng et~al.(2024)Zheng, Li, Liu, Liu, Luan, and
  Wang}]{zheng2024toolrerank}
Yuanhang Zheng, Peng Li, Wei Liu, Yang Liu, Jian Luan, and Bin Wang. 2024.
\newblock Toolrerank: Adaptive and hierarchy-aware reranking for tool
  retrieval.
\newblock \emph{arXiv preprint arXiv:2403.06551}.

\end{thebibliography}
% \clearpage
\appendix

% \section{Related Work}
% % 每个任务有自己的工具。没考虑到同一个任务可能有多个工具，但是这些工具有tradeoff。如何自适应的选择调用哪个？

% \paragraph{Tool Learning}
% Existing tool-learning methods focus on developing different tools to solve various problems. \cite{schick2024toolformer} teaches LLMs to call tools like Calculator, Calendar, etc.
% \cite{qin2023toolllm} facilitates LLMs to master 16000+ real-world APIs.
% \cite{zheng2024toolrerank,qu2024colt,du2024anytool} select the most suitable tool for a given query from a vast tool set, according to the similarity between the query and tool descriptions.
% Differently, we investigate the problem of selecting homogeneous tools, which are less considered. 

% \paragraph{LLMs Selection}
% Earlier, Different LLMs charged differently due to their uneven performance.  Practitioners focus on effectively choosing LLM to save costs.
% \cite{chen2023frugalgpt} gradually employs more expensive LMs until a satisfactory performance is obtained. 
% \cite{lu2023routing,vsakota2024fly,shnitzer2023large} propose the neural routing function that can precisely distribute each query to the expert LLMs.
% % \cite{liu2024optllm} develops a multi-objective optimization method that maximizes the percentage of queries processed accurately, as well as minimizes the total cost of invoking LLMs.
% However, recently, the cost of using LLMs has continued to decrease. 
% The usage cost of tools has become a major obstacle to LLMs-based applications.
% Thus, we explore the selection of homogeneous tools to achieve a balance between performance and cost.

\section{Data Processing Details and Statistics}
\label{app:statistcs}
\paragraph{Construction Details about PrivateTimeQA}
PrivateTimeQA primarily consists of questions related to timeliness. We collected question-answer examples from public QA websites. 
We prioritize privacy protection and data integrity, ensuring that the dataset contains only open-domain QA content and excludes any harmful, or unethical material. 
PrivateTimeQA is used to evaluate LLMs' capability to answer timeliness questions. 
Due to confidentiality requirements, we are unable to publicly disclose the dataset.

\paragraph{Data Processing} 
% FreshQA will be regularly updated by its creator. We have used the version updated at 2024-05-20.
% We use the first 2000 examples in the official HotpotQA testset, and the whole FreshQA datasets as test cases. We randomly select 20000 examples from the official HotpotQA train set as training cases. 
For WebQA and CDQA, we use the official data split.
For each example, we first use different search tools and LLMs for retrieval-augmented generation. We discard examples with invalid search results. After obtaining labels, we merge training examples of CDQA and WebQA, and randomly split the Train/Dev partitions with a ratio of 0.85/0.15. We train the model on the mixed train part and test our model on the testsets of WebQA, PrivateTimeQA, and CDQA.
% Similarly, we train our model on HotpotQA and test it on the testsets of HotpotQA and FreshQA.
The statistics of used datasets are shown in Table \ref{tab:statistcs}.

\begin{table}[!htb]
\centering \small
\begin{tabular}{@{}lllll@{}}
\toprule
 & Language & Train & Dev & Test \\ \midrule
% HotpotQA & English & 16537 & 2918 & 1976 \\
% FreshQA & English & N/A & N/A & 565 \\
WebQA & Chinese & 8102 & 1409 & 2924 \\
PrivateTimeQA & Chinese & N/A & N/A & 1831 \\
CDQA & Chinese & 8067 & 1444 & 1056 \\ \bottomrule
\end{tabular}
\caption{Statistcs of used datasets.}
\label{tab:statistcs}
\end{table}

\paragraph{Training Details}
% For the English experiment, we adopt \textit{FacebookAI/roberta-base} as the backbone. 
We adopt \textit{damo/nlp\_roberta\_backbone\_large\_std} as the backbone. These pre-trained models can be downloaded from the modelscope\footnote{https://www.modelscope.cn/home}.
We use the Adam optimizer and linearly decrease the learning rate to zero with a 10\% warmup ratio. We use PyTorch toolkit to conduct all experiments on the Ubuntu server with a V100 (32G) GPU. All the hyperparameters for are searched according to the final QA accuracy on the development set. For reproduction, we set the random seed to 42 for all experiments. The searched parameters are shown in Table \ref{tab:hyperparameters}.

\begin{table}[!htb]
\centering
\small
\begin{tabular}{@{}lc@{}}
\toprule
batch size & 16 \\
num-epochs & 20 \\
$lr$ & 5e-5 \\
$\alpha$ & 0.8 \\
$k$ & 2 \\ \bottomrule
\end{tabular}
\caption{The used hyperparameters in our experiments.}
\label{tab:hyperparameters}
\end{table}

\end{document}